\documentclass[10pt,twocolumn,letterpaper]{article}

\usepackage{3dv}
\usepackage{times}
\usepackage{epsfig}
\usepackage{graphicx}
\usepackage{amsmath}
\usepackage{amssymb}

\usepackage{epsfig}
\usepackage{pifont}
\newcommand{\cmark}{\ding{51}}
\newcommand{\xmark}{\ding{55}}

\usepackage[pagebackref=true,breaklinks=true,letterpaper=true,colorlinks,bookmarks=false]{hyperref}

\threedvfinalcopy 

\def\threedvPaperID{59} 

\begin{document}

\title{DynOcc: Learning Single-View Depth from Dynamic Occlusion Cues}

\author{
Yifan Wang\textsuperscript{1}\thanks{This work was done while Yifan was an intern at ByteDance.}\qquad 
Linjie Luo\textsuperscript{2}\qquad 
Xiaohui Shen\textsuperscript{2}\qquad 
Xing Mei\textsuperscript{2}\\
\textsuperscript{1}University of Washington\qquad 
\textsuperscript{2}ByteDance AI Lab
}

\maketitle

\begin{abstract}
    Recently, significant progress has been made in single-view depth estimation thanks to increasingly large and diverse depth datasets. However, these datasets are largely limited to specific application domains (e.g. indoor, autonomous driving) or static in-the-wild scenes due to hardware constraints or technical limitations of 3D reconstruction. In this paper, we introduce the first depth dataset \emph{DynOcc} consisting of \emph{dynamic} in-the-wild scenes. Our approach leverages the occlusion cues in these dynamic scenes to infer depth relationships between points of selected video frames. To achieve accurate occlusion detection and depth order estimation, we employ a novel occlusion boundary detection, filtering and thinning scheme followed by a robust foreground/background classification method. In total our DynOcc dataset contains 22M depth pairs out of 91K frames from a diverse set of videos. Using our dataset we achieved state-of-the-art results measured in weighted human disagreement rate (WHDR). We also show that the inferred depth maps trained with DynOcc can preserve sharper depth boundaries.
\end{abstract}


\section{Introduction}

Human visual system can perceive depth from a single view using various monocular cues such as shading, perspective and occlusions. These cues have been used to develop various techniques to teach machines perceive depth information from monocular inputs, such as shape-from-shading~\cite{Zhang1999sfs}, structure-from-vanishing-points~\cite{Criminisi:2002} and structure-from-occlusions~\cite{TomasiK92}. However, these techniques usually pose serious assumptions on the inputs which often lead to mixed degrees of success in practice.

Recently, with the advent of deep learning and affordable depth sensing hardware, significant progress has been made in single-view depth estimation by using the powerful deep learning machinery to harness the increasing amount of available depth data. Many depth datasets emerged, such as NYUDv2~\cite{Silberman2012}, ScanNet~\cite{dai2017scannet}, Make3D~\cite{saxena2009make3d} and KITTI~\cite{Geiger2013IJRR} datasets. Synthetically-generated depth datasets also become available, including SceneNet~\cite{McCormac2016SceneNet} and SUNCG~\cite{song2016semantic}. However, these datasets usually address a specific scenario such as indoor scenes or autonomous driving due to limited hardware capability or acquisition budget. To further improve the performance, more diverse in-the-wild datasets are proposed. Chen et al. \cite{chen2016single} crowdsourced the labeling efforts and created DIW dataset that provides pairwise depth labels for a large number of in-the-wild internet images. Despite its unprecedented data diversity, DIW dataset contains only sparse ground-truth labels, and it is labor-intensive to scale up to more data. Follow-up work therefore resort to using 3D reconstruction techniques to automatically generate 3D geometry ground-truth from in-the-wild images and videos, such as ReDWeb~\cite{Xian_2018_CVPR} using stereo images for disparity maps, YouTube3D~\cite{chen2019learning} using structure-from-motion for sparse point pairs, MegaDepth~\cite{Li2018} and Mannequin~\cite{Li2019} using structure-from-motion and multi-view stereo for dense depth maps. These automatically-reconstructed datasets, due to its technical limitations, can only include \emph{static} scenes. Even if Mannequin~\cite{Li2019} claims that the resulting system can handle dynamic scenes, the fact that it is derived from a special kind of video creation technique still poses limitations on the type of scenes it can cover.

\begin{figure}[t]
    \centering
    \includegraphics[width=\linewidth]{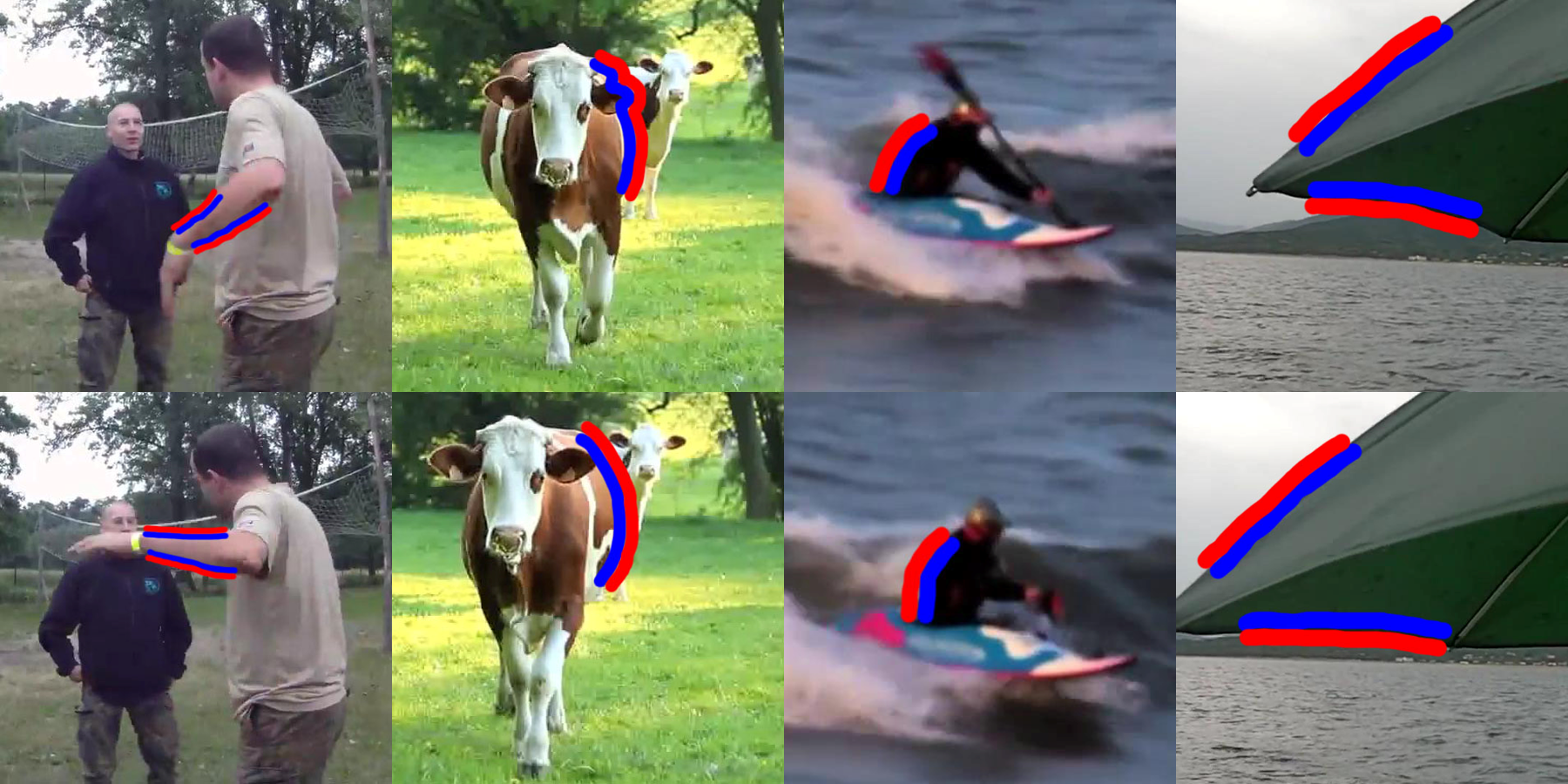}
    \caption{Occlusion cues are prevalent in videos with dynamic scenes, from which pairs with relative depth orders (blue regions are closer than red regions) could be extracted to facilitate the learning of single-view depth estimation.}
    \label{fig:teaser}
\end{figure}

\begin{table*}[t!]
\begin{center}
\begin{tabular}{cccccccccc}
\hline
Dataset       & Scene Type & Dynamic Scene & \# Images & \# Training Pairs/Image\\
\hline
NYUDv2~\cite{Silberman2012}        & indoor &    \xmark      &  795  &   dense    \\
KITTI~\cite{Geiger2013IJRR}         & street &    \xmark      &  93K  &    dense   \\
MegaDepth~\cite{Li2018}     & landmarks &    \xmark   &  130K &    dense   \\
DIW~\cite{chen2016single}           & in-the-wild & \xmark  & 496K  &  1    \\    
Mannequin~\cite{Li2019}     & people &   \xmark      & 170K   &   dense \\
ReDWeb~\cite{Xian_2018_CVPR}        & in-the-wild & \xmark  & 3.6K   &    dense   \\
YouTube3D~\cite{chen2019learning}     & in-the-wild & \xmark  & 795K    & 281   \\
DynOcc (Ours) & in-the-wild & \cmark  & 91K     & 240   \\
\hline
\end{tabular}
\end{center}
\caption{A comparison between existing depth datasets and our DynOcc dataset in terms of scene types, dynamic or static, numbers of images and numbers of pairs per image if available. Our DynOcc dataset contains a large amount of relative depth pairs automatically extracted from dynamic in-the-wild videos.}
\label{tbl:dataset}
\end{table*}

In this paper, we propose \emph{DynOcc}, the first depth dataset that consists of \emph{dynamic} in-the-wild videos. For each video, we provide a number of relative depth pairs for each selected frame in the video. To create these relative depth pairs, we revisit the idea of occlusion cues explored in the previous work~\cite{Sundberg2011} and propose a novel occlusion detection scheme that works well for in-the-wild videos. First, we detect possible occlusion boundaries in a video using two-way optical flow. We then employ a filtering and thinning step that significantly improves the boundary localization accuracy. Next, to robustly determine the occlusion relationship between the two sides of the occlusion boundaries, we move the occlusion boundary by the optical flow vector of each side and check if the boundary is aligned in the next frame; the side with the optical flow vector that aligns the boundary is considered occluding the other side. Finally, we sample in both the occluding and the occluded sides to form depth pairs. In total, we generated 22M depth pairs out of 91K video frames. Our videos are taken from the YouTube-VOS dataset~\cite{xu2018youtube}. A comparison of our depth dataset and other depth datasets is provided in Table~\ref{tbl:dataset}. It should be noted that while we used the YouTube-VOS dataset in this paper, videos with dynamic scenes are much more prevalent than the ones with only static scenes, and our dataset can be easily scaled up with the proposed method.

To evaluate our DynOcc dataset, we follow the training protocol in \cite{chen2016single} using DynOcc and compare the weighted human disagreement rates (WHDR) to various baselines. We achieve state-of-the-art WHDR (10.24\%) by combining both YouTube3D and DynOcc datasets. When we train with  DynOcc, we can still achieve 10.63\% WHDR, which is lower than 10.73\% in our implementation of \cite{chen2019learning} and comparable to 10.59\% WHDR as reported in their paper. These results show that DynOcc is a valuable dataset to improve current state-of-the-art single-view depth estimation systems since it includes more dynamic scenes. We also observe that the inferred depth maps trained with DynOcc preserve sharper depth boundaries thanks to our data distribution more focused on occlusion boundaries. We also perform an ablation study on various depth pair sampling strategies as shown in Table \ref{tbl:ablation}.

To sum up, our contributions are:
\begin{itemize}
    \item The DynOcc dataset, the first depth dataset for dynamic in-the-wild videos.
    \item A state-of-the-art single-view depth estimation system trained with DynOcc dataset.
    \item A robust occlusion boundary detection and foreground-background classification method that works well for dynamic in-the-wild videos.
\end{itemize}

\section{Related Work}

Numerous supervised and unsupervised methods ~\cite{chen2016single,Eigen2014,Fu2018,Godard2017,Laina2016,Li2018,Mahjourian2018,Schonberger2016,Wang2018,Xu2018,Yao2018,Yin2018,Zhou2018,zhou2017unsupervised} have been proposed to estimate dense depth information from a single RGB image or monocular video. A comprehensive survey of different learning mechanisms is beyond the scope of the paper. We mainly review those methods that generate training depth datasets from various sources. 

\paragraph{Depth sensors} A number of RGB-D datasets have been captured with depth sensors, which significantly boost the early research of single view depth estimation methods~\cite{Chang2017,dai2017scannet,Geiger2013IJRR,Geiger2012CVPR,Silberman2012}. However, these datasets are usually captured in indoor scenes with limited depth range~\cite{Chang2017,dai2017scannet,Silberman2012} or from specific outdoor applications such as autonomous driving~\cite{Geiger2013IJRR,Geiger2012CVPR}. Learning from these datasets alone might pose generalization difficulties for real world scenes.

\paragraph{Synthetic data} Another source of RGB-D datasets is synthetic data from realistic rendering techniques~\cite{groueix2018,Lv18eccv,song2016semantic}. These datasets provide high quality RGB-D paired data, which have been shown to be effective for single view tasks such as surface normal prediction~\cite{song2016ssc} and object reconstruction~\cite{Wu2016}. But similar to depth sensors, the diversity of the synthetic data is limited by various factors such as rendering capabilities, asset categories and scene layouts. 

\paragraph{Depth from manual labeling} One common limitation with both depth sensors and synthetic data is that pixel depths are encoded with absolute metric values, which is inherently ambiguous under single view settings~\cite{chen2016single}. Therefore some recent methods propose to use relative depth pairs for single view depth estimation~\cite{chen2016single,chen2019learning,Li2019,Li2018,Xian_2018_CVPR}. Chen \etal first explored this idea by manually labelling relative depth pairs in a large-scale image dataset~\cite{chen2016single}. Their ``Depth in the wild" (DIW) dataset, when combined with existing RGB-D data, significantly improves depth prediction accuracy for real world scenes~\cite{Li2018}. 

\paragraph{Depth from 3D reconstruction} Some recent methods focus on generating relative depth pairs automatically with 3D reconstruction techniques~\cite{chen2019learning,Li2019,Li2018,Xian_2018_CVPR}. Xian \etal~\cite{Xian_2018_CVPR} created the ReDWeb dataset by collecting depth maps from calibrated stereo images; while the MegaDepth dataset proposed by Li \etal~\cite{Li2018} performs Structure-From-Motion (SFM) on online internet photos. Following this line, Chen \etal~\cite{chen2019learning} recently released a large-scale YouTube3D dataset, which extracts relative depth pairs from monocular videos with a quality assessment network and SFM. Compared to stereo images and internet photo collections, monocular videos are more accessible with less capturing bias, but an inherent limitation with SFM is that it can not robustly handle videos with dynamic objects such as human beings. Li \etal~\cite{Li2019} tackled this problem by using SFM on specific internet videos with frozen people and moving cameras. Their Mannequin dataset shows great improvement over human depth prediction in single view images. 

\paragraph{Depth from occlusions} Different from previous methods that rely on 3D reconstruction techniques, we employ the monocular depth cue of dynamic occlusion boundaries. Our method is inspired by the classic figure/ground detection work in~\cite{Sundberg2011}: Sundberg \etal show that optical flow near occlusion boundaries can help determine figure/ground in adjacent regions, which forms a natural data source for relative depth pairs. While this work has been followed up from various perspectives such as depth densification~\cite{Holynski2018}, object segmentation~\cite{Papazoglou2013}, stereo vision~\cite{Wang2019} and light field ~\cite{Tao2013}, its potential has not been fully exploited in single view depth prediction studies. Compared to multi-view reconstruction techniques, dynamic occlusion boundaries can be computed from a much wider range of videos, with fewer restrictions on object and camera movement. We show that relative depth data extracted from occlusion boundaries is a useful complement to existing depth datasets.

\section{Approach}

As shown in Table~\ref{tbl:dataset}, existing training datasets for single-view depth estimation either require great effort to acquire (hardware capture/manual labeling), or are limited to specific scenarios (static scenes). In this work, we propose a method to extract depth occlusion cues from arbitrary videos in the wild, which makes the acquisition of depth training data much easier and more scalable. We show some sample images from our DynOcc dataset in Fig~\ref{fig:dataset}. Our method contains two major steps: occlusion boundary detection and depth order estimation, which we describe in detail in Section~\ref{sec:edge} and Section~\ref{sec:order}, respectively. Based on the estimated occlusion cues, we present our depth pair sampling strategy for training in Section~\ref{sec:sampling}. Finally, the details of generating the dataset are described in Section~\ref{sec:dataset}.

\begin{figure*}[ht]
    \centering
    \includegraphics[width=0.8\linewidth]{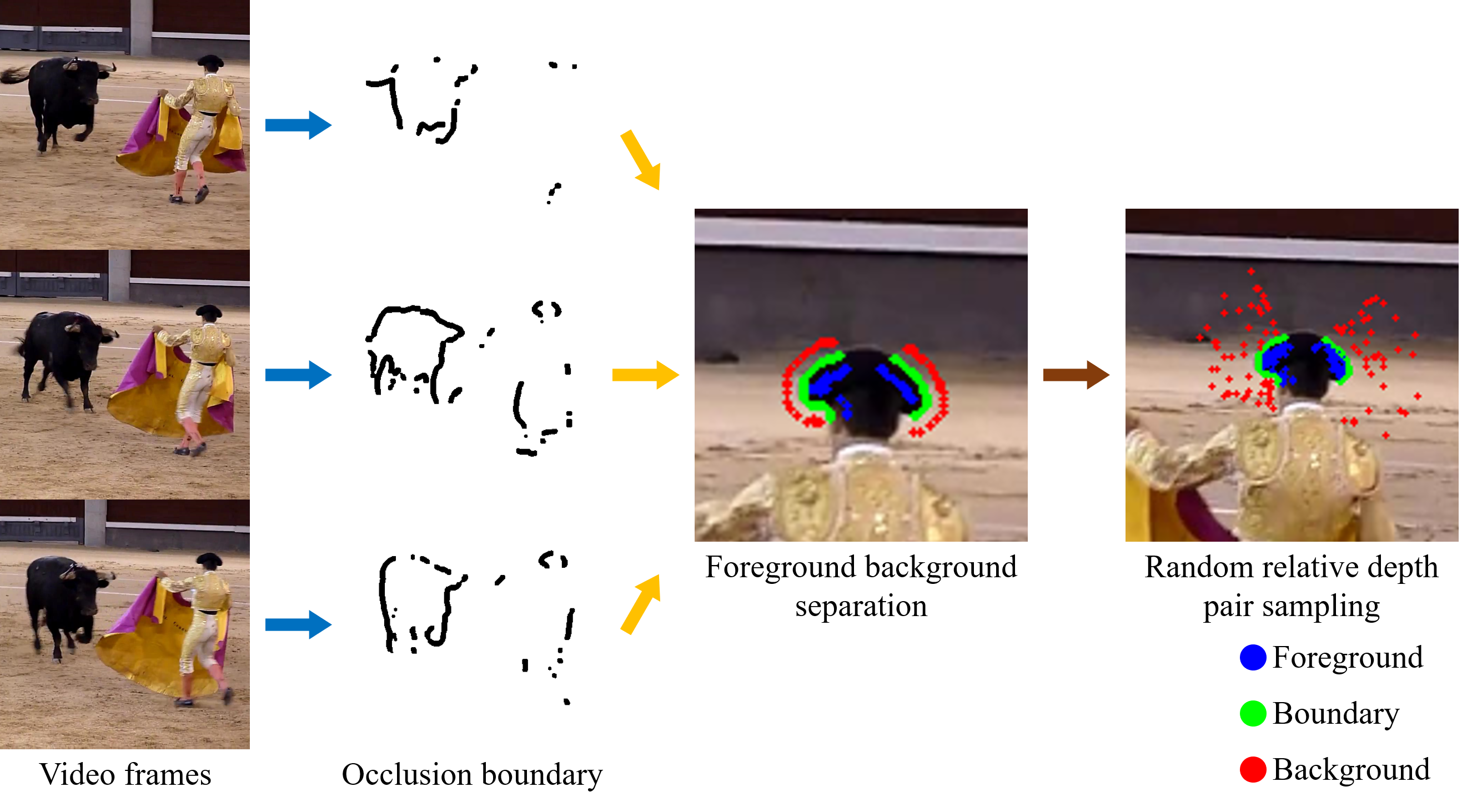}
    \caption{An overview of our data collection method. Our pipeline follows three major steps. Given an arbitrary video, we first extract occlusion boundaries, then separate foreground and background via occlusion boundary matching, finally we randomly sample relative depth pairs from foreground, boundary and background.}
    \label{fig:pipeline}
\end{figure*}
\subsection{Occlusion Boundary Detection}
\label{sec:edge}
Given a video with a dynamic scene, the most reliable relative depth cues would come from the regions near the occlusion boundaries, as the occluding regions next to the boundaries are apparently closer than the occluded regions. Therefore, as the first step of our training data generation pipeline, occlusion boundaries are extracted in each frame. 

There are several prior works that try to extract occlusion boundaries in video frames using optical flow~\cite{Sundberg2011,Holynski2018}. In particular, Holynski and Kopf~\cite{Holynski2018} proposed a two-step approach to extract occlusion boundaries for depth densification, where a soft depth edge map is generated by calculating flow gradients, from which exact depth edges are then extracted and connected using Canny edge detector. While the extracted edges are mostly clean and complete in their work, we observed that some text edges are misclassified as depth edges as well due to the inclusion of Canny edges, which will yield incorrect relative depth pairs and largely affect the training of the depth estimation network.

To ensure the precision of the extracted depth edges, we chose not to use the Canny edges, and instead largely follow the first step in \cite{Holynski2018} to obtain the soft depth edge map, and then keep the most confident ones with edge thinning and thresholding. 

In particular, we use FlowNet 2.0 \cite{IMKDB17} to compute a dense flow field, and identify the regions with large flow magnitude $|| \nabla F ||_1$, since large changes in the flow correspond to depth discontinuities because of parallax. As observed in \cite{Holynski2018}, optical flow is usually not reliable near converging occlusion boundaries, where the flow directions of the two sides next to the boundary are converging to each other, and the pixels on the occluded side are not visible in the nearby frame. On contrary, the flow near the diverging boundaries are more reliable, as the occluding region is leaving the occluded one, and pixels on both side are visible in the nearby frame.

Therefore, we compute two flow fields $F_{prev}$ and $F_{next}$ w.r.t. two nearby frames, one backward and one forward, and only retain the diverging edges of the two flow fields as our occlusion boundary candidates. Specifically, we calculate a occlusion boundary confidence map $B$ for each pixel $p$. Given a pixel of interest $p$, we find two helper pixels $h_1$ and $h_2$ that are offset at unit distance in the gradient direction $d$ and its opposite. We compute the projection of the flow vectors at $h_1$ and $h_2$ on gradient direction $d$: 
\begin{equation}
    f_1 = F[h_1] \cdot d, f_2 = F[h_2] \cdot d
\end{equation}
and take their difference as the confidence score of $p$,
\begin{equation}
    b = f_2 - f_1
\end{equation}
Evidently, $b$ will be positive (reliable) for diverging flow projections, and negative (unreliable) for converging flow projections.

We then fuse the gradient magnitude of the two flow fields into the occlusion boundary confidence map $B$ by selecting the more reliable quantity at each pixel $p$:
\begin{equation}
    B[p] = 
    \begin{cases}
       || \nabla F_{prev} [p] ||_1 , & b_{prev}[p] > b_{next}[p]\\
       || \nabla F_{next} [p] ||_1,  & b_{prev}[p] \leq b_{next}[p]
    \end{cases}
\end{equation}

After computing the confidence map $B$, we blur it with a wide box filter of size $k = 31$ to connect short edge segments. We also normalize $B$ by dividing the $90^{th}$ percentile value. This makes the parameter settings more invariant to the video content. We then threshold the confidence map to only keep the most confident edges\footnote{We set the threshold to be 0.3, which we empirically found can keep most correct depth edges.}. The final occlusion boundaries are then extracted using the thinning method in \cite{guo1989parallel}.

It should be noted that the extracted edges are not complete occlusion boundaries of the objects, but rather some edge segments, as shown in Fig.\ref{fig:pipeline} and Fig.\ref{fig:dataset}. However, it is not necessary to obtain all the occlusion boundaries in our work, as long as there are sufficient edge segments from which we can generate correct training pairs.  

Some of the extracted edges are still not reliable, which we further prune in the subsequent depth Order estimation step using a validation approach.

\subsection{Depth Order Estimation}
\label{sec:order}
After extracting the occlusion boundaries, we need to decide the depth order around the boundaries, i.e., which side of the edge is closer to the camera. It is equivalent to finding which side is occluding the other one, as the occluding region is probably closer than the occluded ones. As discussed in \cite{Sundberg2011}, the occlusion boundary moves the
same way as the occluding region next to it. Therefore we can assign the occluding/occluded regions by comparing their motion with the motion on the occlusion edge.

The motion on the occlusion boundaries calculated by FlowNet is not reliable. Therefore instead of directly using the estimated flow on the edges, we take an assumption-validation approach, that is, we first assume one side of the edge is the occluding region (foreground), and assign its flow vector to the edge. We then move the occlusion boundary by the assigned flow vector and check if the boundary is aligned in the next frame.

\begin{figure}[t]
    \centering
    \includegraphics[width=\linewidth]{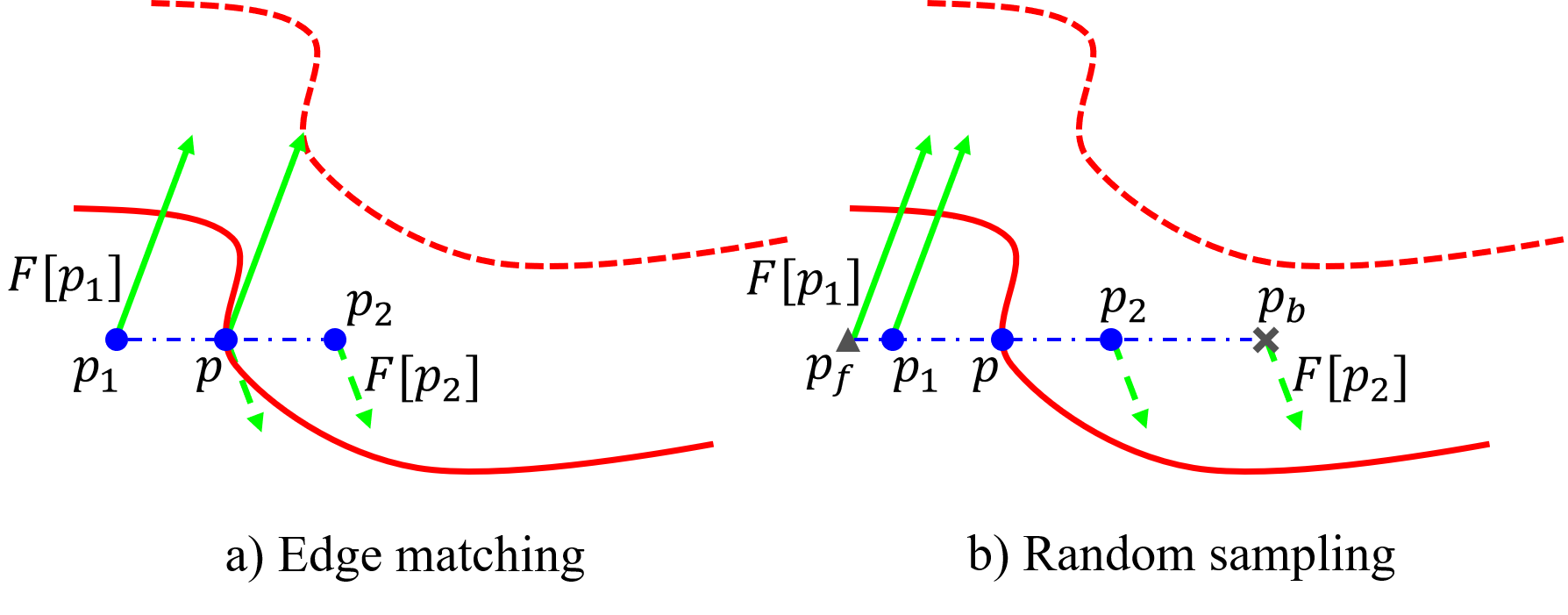}
    \caption{An illustration of our depth order estimation and relative depth pair sampling algorithm. For each pixel $p$ on the occlusion boundaries segment, we find two helper pixels, $p_1$ and $p_2$, along the gradient direction. We then try to warp $p$ using each pixel's flow $F[p_1]$ and $F[p_2]$. On the left, $p$ is warped to the occlusion boundary (dashed red line) in the nearby frame using $F[p_1]$. Therefore, we consider $p_1$ as foreground pixel and $p_2$ background. On the right, we randomly sample foreground and background pixels $p_f$ and $p_b$ along flow direction $F[p_1]$ and $F[p_2]$ respectively. We take $(p_f, p), 0$ and $(p, p_b), +1$ as the resulting relative depth pairs.}
    \label{fig:detail}
\end{figure}

To be specific, we first split occlusion boundary into short segments $s$, each with $c$ pixels. For each pixel $p$ in segment $s$, we find its gradient direction $d$ (blue dotted line in Figure~\ref{fig:detail}). Along the direction $d$, we sample two helper pixels $p_1$ and $p_2$ at a small offset on each side.
If we assume the region including $p_1$ is the foreground, the flow $F$ at $p_1$ is assigned to pixel $p$. We can then move pixel $p$ using the flow $F[p_1]$ to the nearby frames. If the flow is correct, the warped pixel should be aligned with a pixel on the occlusion edges in the other frame. Ideally, only one of the flows $F[p_1]$ and $F[p_2]$ is correct, and we can identify the foreground accordingly. In some cases where the foreground and the background move in the same direction but with different magnitude, pixel $p$ could be aligned using both $p_1$ and $p_2$'s flow. To solve such ambiguities, we perform the same procedure for all the pixels in edge segment $s$, and count the aligned pixels $c_1$ and $c_2$ for both sides. We say the side of $p_1$ is matched to be foreground via flow $F$ if
\begin{equation}
(c_1 - c_2) / c  > \delta
\end{equation}
where $\delta$ is a pre-defined threshold to filter out those unreliable edges. Similarly, the side including $p_2$ is considered closer if $(c_2 - c_1) / c  > \delta$.

To add more robustness against the noise in flow estimation and depth edge extraction, we only assign the side of $p_1$ to be foreground when the side of $p_1$ is matched to be foreground via both $F_{prev}$ and $F_{next}$.

\subsection{Relative Depth Pair Extraction}
\label{sec:sampling}
After determining the depth order near the occlusion edge segment $s$, we need to extract relative depth pairs for training. Relative depth pair $(i, j), o$ consists of a pair of points $(i, j)$ and its ordinal relation $o \in \{-1, 0, 1\}$ ($o = +1$: $i$ is closer, $o = -1$: $j$ is closer, and $o = 0$: $i, j$ have the same depth). One simple way to do this is to use all the pairs of $(p_1, p_2), +1$ sampled in Section~\ref{sec:order} when estimating depth order. However, this sets a fixed distance between all $p_1$s and $p_2$s, which would induce a bias in the training data and lead to the halo effect in the network output, as shown in Figure~\ref{fig:halo}. 

To fix this issue, we adopt a random sampling strategy. Take the the background pixel $p_b$ for example, we sample one pixel along the direction from $p$ to $p_2$ with a random offset. To ensure the sample pixel is still on the background, we check if $p_2$ and $p_b$ share the same flow vector, and repeat the sampling process until the criterion is satisfied. Similar strategy is adopted for foreground pixel sampling. We found that random sampling can largely reduced the halo effect, and the results are not sensitive to the value of random offset, as long as the random sampling strategy is enabled.

However, the network trained with those sampled pairs would still produce depth discontinuities on many texture edges. That is because occlusion edges usually come with sudden texture changes. Training with the relative pairs purposed above would make the network focus on the texture changes instead of perceiving depth boundaries. To alleviate the problem, we additionally sample negative examples, i.e., pairs with similar depths ($o=0$), during training. 

We use the same sampling strategy to search along $p_f$'s direction and find a pixel $p_l$ that shares the same flow vector with $p_f$ (as well as $p$). $p$ and $p_l$ belong to the same foreground region, and should have similar depth. Therefore we take pair $(p, p_1)$ as an additional training sample with ordinal relation $o=0$.
 
Thus for each pixel $p$ on the occlusion edge segment, we extract two relative depth pairs: one positive sample and one negative sample. Even though we have eliminated many unreliable depth edges, it could still generate a gigantic number of training samples if we sample all the possible pairs from the remaining depth edges. The depth order information along the same depth edge has a lot of redundancy. At the same time, having too many training samples would make the training of the networks much slower. Therefore it is neither necessary nor efficient to extract all the relative pairs. We chose to randomly pick a smaller number of the total extracted relative depth pairs as supervision during train time. In the experiments we found that the results are quite robust against different random sampling rates, and we chose 10\% in the experiment for training efficiency.  

\begin{figure*}[t!]
    \centering
    \includegraphics[width=\linewidth]{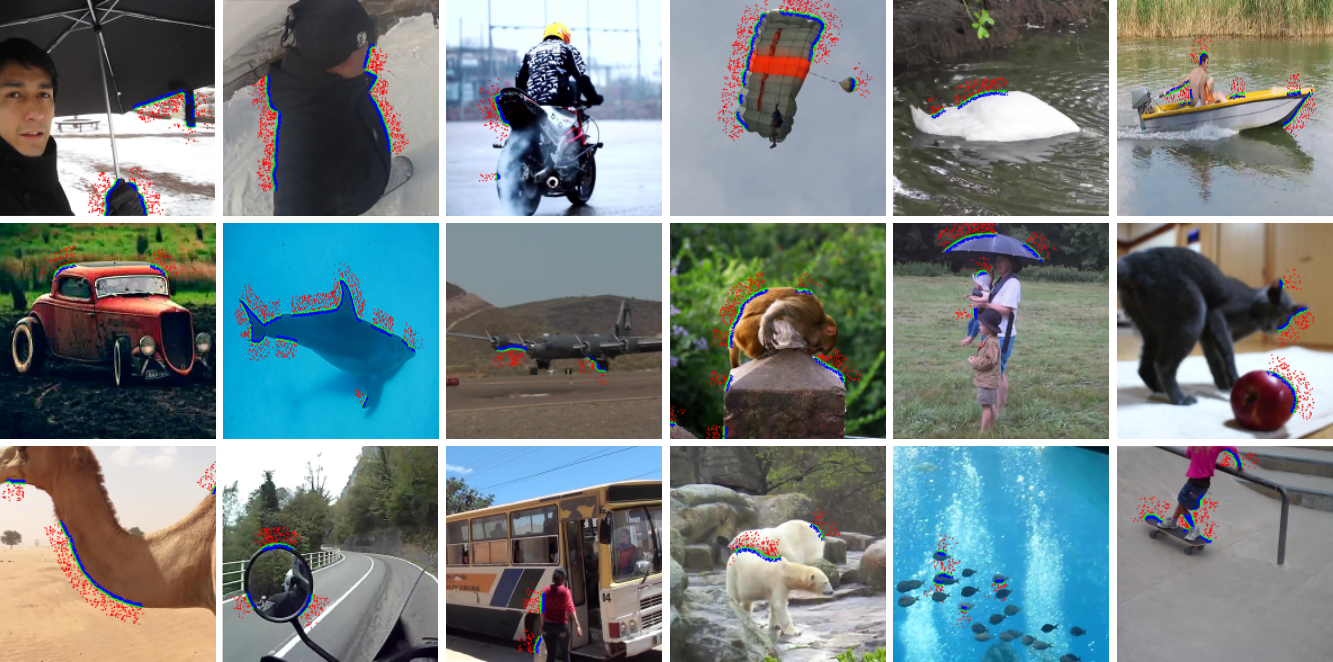}
    \caption{Examples of automatically collected relative depth annotations in DynOcc. The blue points are in the foreground, green points on the boundary and red points in the background. The relative relation is that green and blue points should have the same depth while green points should be closer than the red ones. These relative depth pairs are mostly correct.}
    \label{fig:dataset}
\end{figure*}

\subsection{Implementation Details}
\label{sec:dataset}
We use video frames from YouTube-VOS \cite{xu2018youtube} as our original source videos. It contains 4,453 YouTube video clips covering various scene and object types. We detect occlusion boundaries for every frame and align edges with nearby frame with a baseline of 2 frames. We set two help pixels $p_1$ and $p_2$ to be 5 pixels away from $p$. 

The foreground region tends to have more thin structures such as tree branches and limbs, as well as sudden depth changes within its region. Therefore when sampling pair $p_f$ and $p_b$, we randomly sample $p_b$ within 30 pixels from the occlusion edge, and sample $p_f$ within 7 pixels away.
By having $p_f$ closer to $p$, it is less likely to has sudden depth changes within such a small region. Therefore, we can consider these two points have the same depth.

Finally, after applying our method on YouTube-VOS, we obtain 90,905 images, with an average of 240 relative depth pairs per image. Example images and annotations of DynOcc are shown in Figure ~\ref{fig:dataset}. As we can see, the dataset covers a diverse range of scene and object types in-the-wild, and can provide accurate depth cues in many challenging cases, which are not available in previous datasets, e.g., the depth boundaries between the two polar bears in the last row.

\section{Experiments}

\begin{figure*}[ht!]
    \centering
    \includegraphics[width=\linewidth]{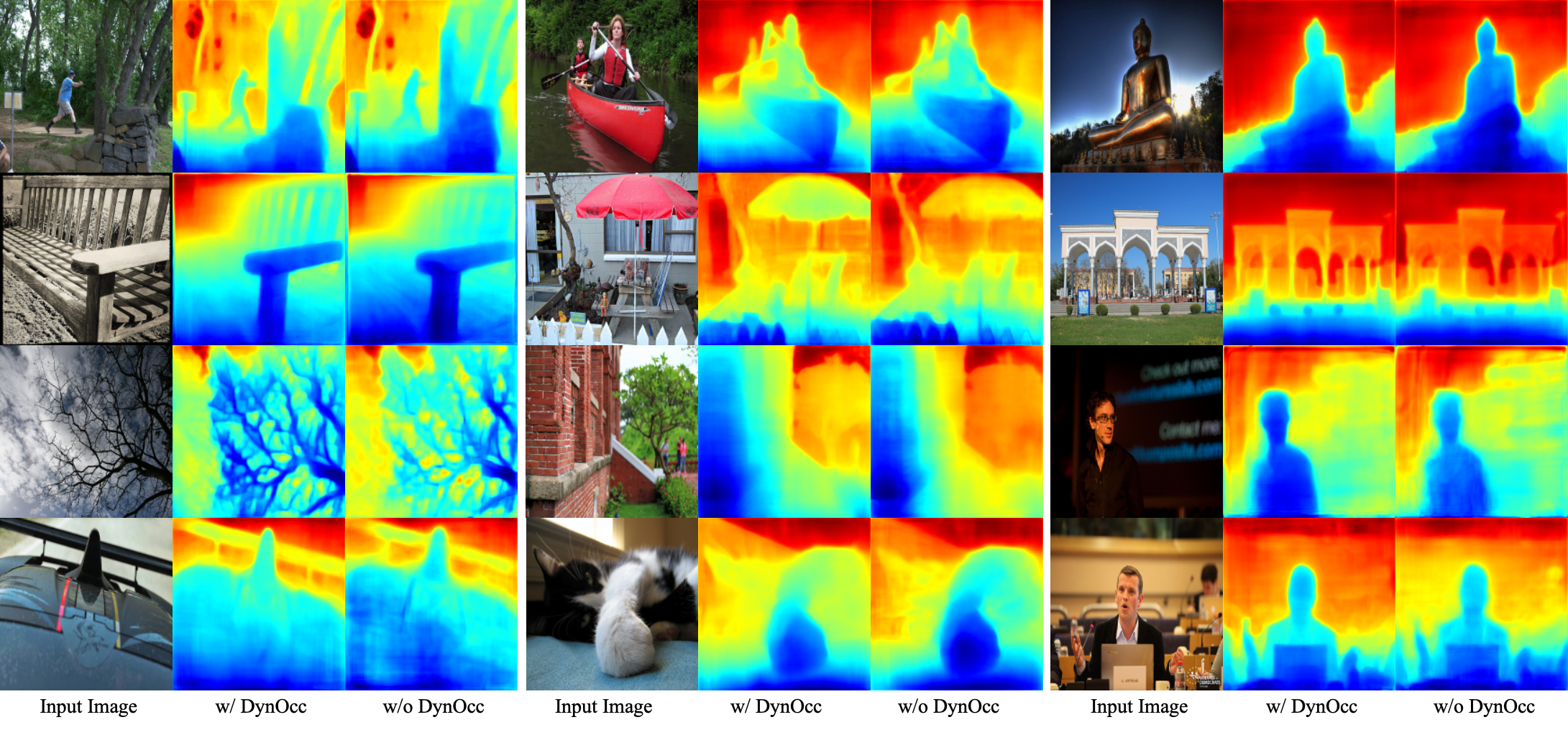}
    \caption{Qualitative results on the DIW test set by the EncDecResNet \cite{Xian_2018_CVPR} trained on ImageNet + ReDWeb + DIW + YouTube3D (w/o DynOcc), and trained with DynOcc (w/ DynOcc).}
    \label{fig:results}
\end{figure*}

\subsection{Single-View Depth Estimation}
\label{sec:training}
We use the state-of-the-art EncDecResNet \cite{Xian_2018_CVPR}, which is an encoder-decoder network based on ResNet50, as our network architecture. To use relative depth pairs as supervision, we use the same ranking loss as in \cite{chen2016single}. Given a training image $I$ and its $K$ queries $R = \{(i_k, j_k, o_k)\}$, $k = 1, \cdots, K$, where $i_k$ and $j_k$ are the two different query points respectively, and $o_k\in \{+1, -1, 0\}$ is the ground truth ordinal relation between $i_k$ and $j_k$: closer ($+1$), further ($-1$) and almost the same ($0$). Let $z$ be the predicted depth map and $z_{i_k},z_{j_k}$ be the predicted depth at $i_k, j_k$. The loss function is 
\begin{equation}
    L(I, R, z) = \sum_{k=1}^K \psi (I, i_k, j_k, r, z)
\end{equation}
where $\psi (I, i_k, j_k, r, z)$ is the loss function for the $k$-th query
\begin{equation}
    \psi (I, i_k, j_k, r, z) = 
    \begin{cases}
       \log ( 1 + \exp (-z_{i_k} + z_{j_k})), &o_k = +1\\
       \log ( 1 + \exp (z_{i_k} - z_{j_k})), &o_k = -1\\
       (z_{i_k} - z_{j_k})^2, &o_k = 0
    \end{cases}
\end{equation}
This ranking loss encourages a small difference between depths if the ground-truth relation is almost the same ($o=0$); otherwise it encourages a large difference.

We follow the same training paradigm as in \cite{chen2019learning}. Following \cite{Xian_2018_CVPR}, we only back propagate the loss for top $75\%$ of the queries with the largest loss. Moreover, we pretrain our encoder ResNet 50 on ImageNet. 

Note that the datasets for single-view depth estimation are collected in different ways and are complementary to each other. As shown in \cite{chen2019learning}, combining different datasets for training usually yield the best results. Therefore instead of compare the performance using each dataset individually, following \cite{chen2019learning}, we would like to evaluate if our new DynOcc dataset is valuable by measuring if it can further improve the performance of single-view depth estimation when used together with other datasets.

Besides DynOcc, We used the following datasets for network training: DIW~\cite{chen2016single}, ReDWeb~\cite{Xian_2018_CVPR}, YouTube3D~\cite{chen2019learning}. We measure the performance of the trained network on DIW test set \cite{chen2016single} by the weighted human disagreement rate (WHDR), i.e. the percentage of incorrectly ordered point pairs.

\begin{table}
\begin{center}
\begin{tabular}{c c c c | c}
\hline
ReDWeb & DIW & YouTube3D & DynOcc & WHDR \\
\hline
\cmark & \cmark &  &  & 10.95\% \\
\cmark & \cmark & \cmark &  & 10.73\%\\
\hline
\cmark & \cmark &  & \cmark & 10.63\% \\
\cmark & \cmark & \cmark & \cmark & \textbf{10.24}\% \\
\hline
\end{tabular}
\end{center}
\caption{All models are using EncDecResNet \cite{Xian_2018_CVPR} and the encoder is pretrained on ImageNet.}
\label{tbl:results}
\end{table}

\subsection{Comparison with State-of-the-art}
We compare our dataset with other recently proposed relative depth dataset: DIW \cite{chen2016single}, ReDWeb \cite{Xian_2018_CVPR}, and YouTube3D \cite{chen2019learning}. We follow the recent trend of mixed training~\cite{chen2019learning} to get the best result. We show quantitative result in Table~\ref{tbl:results}. Since \cite{Xian_2018_CVPR} did not provide implementation, we implement EncDecResNet based on their description in the paper. 

As a validation of our implementation, we first test our implementation on DIW + ReDWeb. Notice that DIW has much more images than ReDWeb (496K vs 3.6K), we add a multiplier of 30 to ReDWeb when sampling training examples, and combine it with DIW. Our model achieves an error rate of 10.95\% on DIW test set, which is slightly better the 11.37\% reported in \cite{Xian_2018_CVPR}. We then use this setting as a baseline and compare DynOcc with the state-of-the-art YouTube3D. 

We then add YouTube3D to the DIW + ReDWeb for joint training. This achieves an error rate of 10.73\% in our implementation, compared to 10.59\% as reported in \cite{chen2019learning}. This number is the current state of the art on DIW.

As a direct comparison against YouTube3D, we add DynOcc to the DIW + ReDWeb. Similar to the training paradigm in \cite{chen2019learning}, we first train our model on ReDWeb and DynOcc for 20 epochs using a learning rate of $10^{-4}$. Then we fine tune our model on DIW for 10 epochs using a learning rate of $10^{-5}$. Our model achieves an error rate of 10.63\%, which is comparable to the 10.73\% given by YouTube3D. This suggests that DynOcc is useful in further improving single-view depth estimation. This is noteworthy because YouTube3D has much more training pairs than DynOcc (223M vs 21M). Using DynOcc as a supplementary dataset, EncDecResNet can achieve better results with much less supervision.

Finally, we train an EncDecResNet on the combination of DIW, ReDWeb, YouTube3D, and DynOcc. We first train the EncDecResNet on DIW, ReDWeb, YouTube3D and DynOcc for 20 epochs using a learning rate of $10^{-5}$. Then we fine tune our model on DIW for 10 epochs using a learning rate of $10^{-5}$. Our model achieves an error rate of 10.24\%, a new state of the art performance on DIW. This result demonstrates the effectiveness of DynOcc as a supplementary single-view training data.

We show qualitative results in Figure~\ref{fig:results}. We compare our full model with the current state of the art on DIW (trained with DIW+ReDWeb+YouTube3D). By adding DynOcc, the model performs better on depth edge regions. Our results have sharper edges and can predict the depth of thin structures more accurately. The planar surfaces are also more smooth in our results due to the added negative samples.

The above results suggest that our proposed method can extract high-quality relative depth pairs from dynamic videos with high accuracy. Such results are significant, because our dataset is gathered by a completely automatic method. Our automatic method can be readily applied to a much larger set of Internet videos and thus has potential to advance the state of the art of single-view depth even more significantly.

\begin{figure}[t!]
    \centering
    \includegraphics[width=\linewidth]{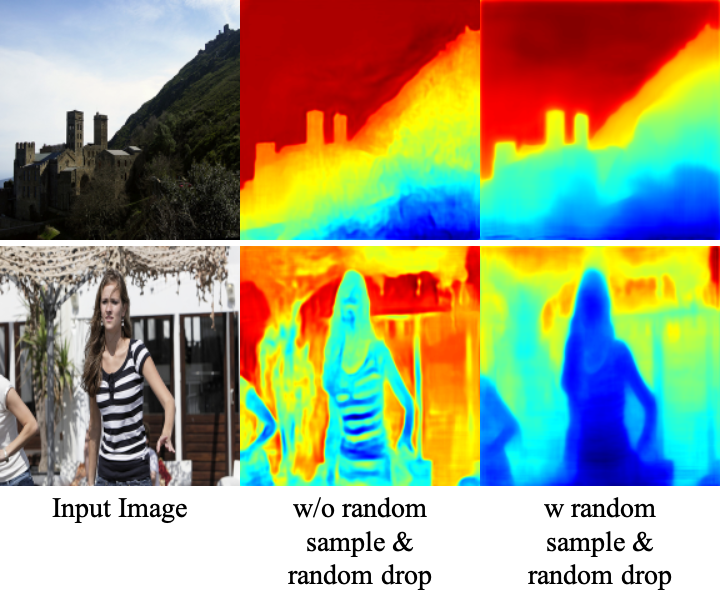}
    \caption{Results using DynOcc without random sampling and random drop. The depth map is heavily affected by texture, and has halo effect around boundaries.}
    \label{fig:halo}
\end{figure}



\begin{table}[t!]
\begin{center}
\begin{tabular}{c c | c}
\hline
Random distance & Random drop & WHDR \\
\hline
\xmark & \xmark & 12.28\% \\
\cmark & \xmark & 11.06\% \\
\cmark & \cmark & 10.63\% \\
\hline
\end{tabular}
\end{center}
\caption{Ablation study is done on DIW + ReDWeb + VOS. All models are using EncDecResNet \cite{Xian_2018_CVPR} and the encoder is pretrained on ImageNet.}
\label{tbl:ablation}
\end{table}

\begin{table}[t!]
\centering
\begin{tabular}{c|cccc}
Drop rate & 1/5 & 1/10  & 1/15  & 1/20  \\
\hline
WHDR & 10.45\% & 10.24\% & 10.26\% & 10.27 \% 
\end{tabular}
\caption{Ablation study is done on DIW + ReDWeb + VOS + YouTube3D. All models are using EncDecResNet \cite{Xian_2018_CVPR} and the encoder is pretrained on ImageNet. The drop rate has a small impact on WHDR.}
\label{tbl:drop_rate}
\end{table}

\begin{figure}[htbp!]
    \centering
    \includegraphics[width=\linewidth]{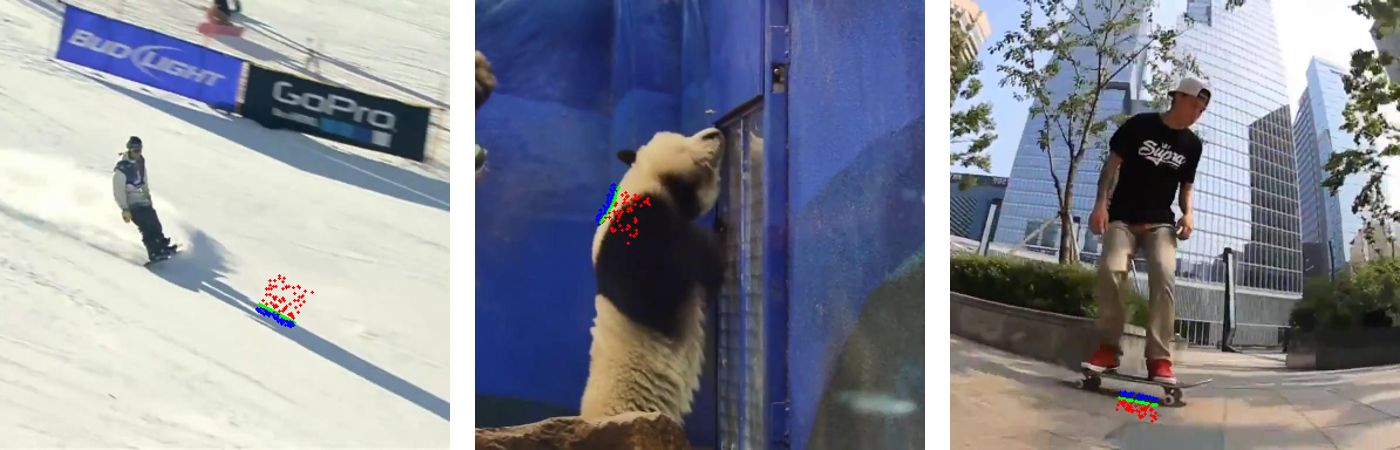}
    \caption{Failure cases in DynOcc. Our algorithm does not perform well on shadows. Sometimes the flow result is inaccurate, leading towards mismatch.}
    \label{fig:failure}
\end{figure}

\subsection{Ablation Study}
As ablation study, we test EncDecResNet on three variants of our relative depth pair extraction method: first, only using the fixed $(p_1, p_2)$ pairs as supervision; second, using the sampling strategy in Section~\ref{sec:sampling} along the gradient direction, and using all the sampled point pairs, without random drop; finally, using both the sampling strategy in Section~\ref{sec:sampling} and picking only $10\%$ of all the pairs.

We train all three models on DIW, ReDWeb and DynOcc for 20 epochs using a learning rate of $10^{-5}$, and then fine tune them on DIW for 1 epoch using a learning rate of $10^{-5}$.

We show quantitative results in Table~\ref{tbl:ablation} and qualitative results in Figure~\ref{fig:halo}. The numbers prove the effectiveness of adding random samples and random drop. They prevent the network from focusing on texture edges and producing result with halo effects. Furthermore, it also helps the network produce more smooth depth in the planar regions due to the addition of negative samples. In Table~\ref{tbl:drop_rate} we show that the performance is not sensitive to the portion of samples used in the training.

Finally, we show some misclassified samples in our DynOcc dataset in Figure~\ref{fig:failure}. They are usually caused by inaccurate optical flow estimation, or some ambiguous cases, e.g., the shadows in the first and third image, which have the same motion as the persons but are in the background areas. Since the percentage of such failure cases is very small, their effect on the overall training is very limited. The dataset can be further refined with better flow estimation or shadow detection, which we leave as future work.

\section{Conclusion}

In this paper, we introduce DynOcc, the first depth dataset that consists of dynamic in-the-wild videos by leveraging the occlusion cues in the videos. We propose a novel occlusion detection and depth order assignment method that works accurately for in-the-wild videos. We show that we can achieve state-of-the-art weighted human disagreement rates (WHDR) in single-view depth estimation task using DynOcc thanks to its more dynamic data distribution. We also show that our inferred depth maps preserve sharper depth boundaries for a variety of input images.

\newpage

{\small
\bibliographystyle{ieee}
\bibliography{depth}
}

\end{document}